\documentclass[10pt,journal,compsoc]{IEEEtran}
\ifCLASSOPTIONcompsoc
  \usepackage[nocompress]{cite}
\else
  \usepackage{cite}
\fi
\ifCLASSINFOpdf
  \usepackage[pdftex]{graphicx}
  \graphicspath{{./figures/}{./figures_icfhr/}}
\else
\fi

\usepackage{amsmath}
\usepackage{url}

\usepackage{xcolor}

\makeatletter
\newcommand{\thickhline}{%
	\noalign {\ifnum 0=`}\fi \hrule height 1pt
	\futurelet \reserved@a \@xhline
}
\makeatother

\newcommand{\thickcline}[1]{%
	\@thickcline #1\@nil%
}

\usepackage{amssymb}
\usepackage{booktabs}

\begin{document}
%
\title{Neural Word Search in Historical Manuscript Collections}

%

\author{Tomas~Wilkinson, Jonas~Lindstr{\"o}m, and Anders~Brun
\IEEEcompsocitemizethanks{
\IEEEcompsocthanksitem Tomas Wilkinson and Anders Brun are with the Department of Information Technology, Uppsala University, Sweden.\protect\\
E-mail: \{tomas.wilkinson, anders.brun\}@it.uu.se
\IEEEcompsocthanksitem Jonas Lindstr{\"o}m is with the Department of History, Uppsala University, Sweden.\protect\\
E-mail: jonas.lindstrom@hist.uu.se
\IEEEcompsocthanksitem A preliminary version of this paper \cite{wilkinson2017neural} appeared in ICCV 2017}
}

\IEEEtitleabstractindextext{%
\begin{abstract}
We address the problem of segmenting and retrieving word images in collections of historical manuscripts given a text query. This is commonly referred to as "word spotting". To this end, we first propose a model based on deep neural networks that we dub Ctrl-F-Net. The model simultaneously generates region proposals and embeds them into a word embedding space, wherein a search is performed. We further introduce a simplified version called Ctrl-F-Mini. It is faster with similar performance, though it is limited to more easily segmented manuscripts. We evaluate both models on common benchmark datasets and surpass the previous state of the art. Finally, in collaboration with historians, we employ the Ctrl-F-Net to search within a large manuscript collection of over 100 thousand pages with varying styles, written across two centuries. With only 11 training pages, we enable large scale data collection in manuscript-based historical research. This results in a speed up of data collection and the number of manuscripts processed by orders of magnitude. Given the time consuming manual work required to study old manuscripts in the humanities, quick and robust tools for word spotting has the potential to revolutionise domains like history, religion and language. 
\end{abstract}

}

\maketitle

\IEEEdisplaynontitleabstractindextext

%
\IEEEpeerreviewmaketitle

\IEEEraisesectionheading{\section{Introduction}\label{sec:introduction}} 

\IEEEPARstart{S}{ince} the invention of writing over 5000 years ago, people have been recording a great variety of things in written form. Up until the 20\textsuperscript{th} century, writing has been the primary way of storing information across generations. Consequently, manuscripts are a vital source of knowledge about the past, and for that reason of great interest to contemporary historical research. Once digitized, manuscripts are typically collected in archives \cite{vaticanLibrary, old_bailey, GenderAndWork}. Some are also transcribed \cite{old_bailey}, allowing for text-based searching and processing \cite{Hitchcock_Turkel_2016}. Others are collections of published data used in research \cite{GenderAndWork, pettersson2016histsearch}, though they are most commonly simply photographed \cite{vaticanLibrary}. Research in historical documents often consists of manually searching for small and scattered pieces of information in large amounts of digitized manuscripts. Finding where to look, or even which book to examine can be time consuming, as it is commonplace for a researcher to spend several months with a single book of a few hundred pages. It is a matter of looking for a black cat in a coal cellar. A need for a scalable solution exists as large part of the writings produced throughout history are yet to be studied.

\begin{figure}[t!]
	\begin{center}
	\includegraphics[width=\linewidth]{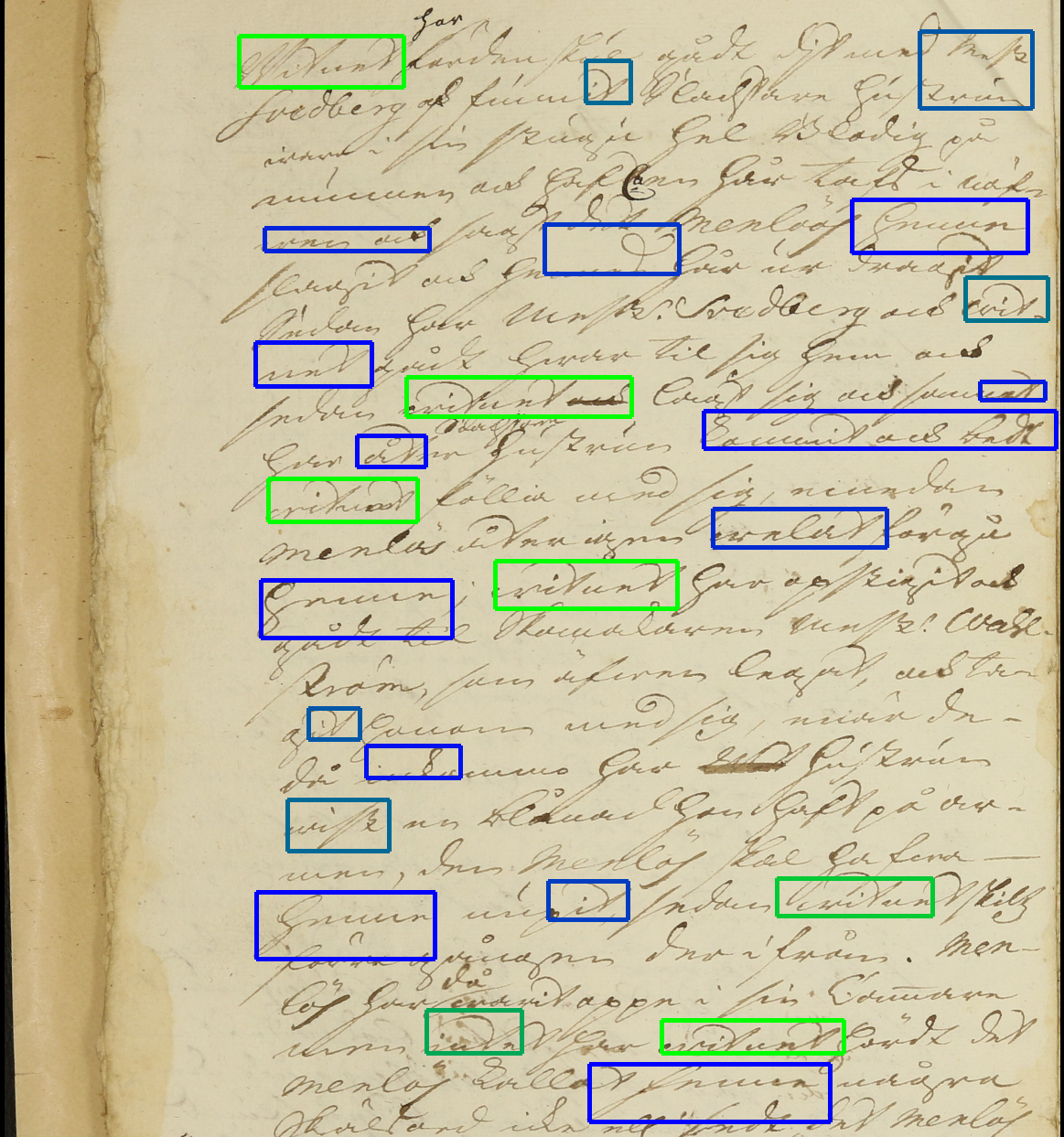}
	\end{center}
	\caption{A search for the word "witnet" (eng. "the witness") on one page of the Snevringe court records dataset, displaying the top 25 results. The greener the bounding box, the closer it is to the search query. Note that the true number of occurrences of the word "witnet" is 6 and that many of the other results are parts of words that are similar to the query.}
	\label{fig:example_search}
\end{figure}

Word spotting \cite{WordSpotting1996, giotis2017survey} is a way to address the problem of finding where to look. The task consists of locating and retrieving images of words given a user supplied query, much like a regular text word search found in common text processors. 

Since its introduction, two principal modes of variation on the task have been proposed. The first is input data format, where images can either be of full manuscript page, known as \emph{segmentation-free} word spotting; or pre-cropped individual word images, known as \emph{segmentation-based} word spotting. The second variation is the type of query. It may either be a cropped word image, also called \emph{Query-by-Example} (QbE); or a text string, also called \emph{Query-by-String} (QbS) or out-of-vocabulary word spotting. All else being equal, QbS is preferred over QbE as the user is not required to find an instance of the query before being able to search. In real-life practical settings, segmentation-based word spotting is not applicable. First, because the automatic segmentation of a manuscript into words is non-trivial in most handwritten documents. Second, because the manual work of segmenting individual words is very time consuming, almost to the same extent as transcribing a manuscript.

An alternative to word spotting that has been successfully applied to manuscripts is crowdsourcing the transcription of a collection of manuscripts \cite{causer2012building, oldweather, AmazonTurk2011, BH2M2014}. Crowdsourcing typically provides a good quality transcription of a source material, enabling the use of text-based search and analysis tools. Nevertheless, there are some drawbacks with crowdsourcing, in particular with regards to historical manuscripts. They are often written in esoteric languages that require specialized skills to transcribe. Another drawback is a possible difficulty in attracting volunteers when working with unknown, prosaic source material lacking the fame, prestige, and historical importance of successful crowdsourcing projects \cite{causer2012building, oldweather, BH2M2014}. Furthermore, it does not scale well compared to word spotting, and requires a priori identification of interesting manuscripts to select for a crowdsourcing project. 

Another related technology that naively one might try, but falls short, is optical character recognition (OCR) \cite{MoriOCR1992}. As OCR was developed primarily for reading machine printed text, it has many limiting assumptions that are unrealistic when it comes to historical handwritten manuscripts. They include easily separated letters, minor writing style variation, and a canonical orthography, all of which limit what manuscripts are possible to study and are not necessarily true for historical manuscripts.

Some of the shortcomings of OCR technologies have been addressed with Handwritten Text Recognition (HTR), where many of the assumptions of OCR software have either been relaxed, or removed altogether. Much work has been done on HTR on handwritten manuscripts \cite{marti2001using, graves2009novel, pham2014dropout, bluche2017scan}, though they come with their own set of assumptions and restrictions. First, compared to word spotting you typically need a relatively large amount of training data to learn a model good enough to perform accurate recognition. Second, current HTR methods typically take text lines as input, requiring an initial text line segmentation step that potentially introduces uncorrectable errors. Considering the messy and compressed layout of many historical manuscripts, segmenting the lines is a challenging task. Third, HTR methods typically rely heavily on language models, which in turn require a large amount of training data that might not be available in the uncommon languages that many manuscripts are written in. Moreover, for the application detailed in the paper, getting a line-by-line transcription is not desirable, as the methodology is based on searching for certain keywords and then reading and interpreting based on the surrounding context. In this case, transcribing the text would only be done for the possibility of searching for keywords, which is what word spotting does directly. 

\subsection{Contributions}
The contributions of this paper include:

\begin{enumerate}
	\item Two models for segmentation-free query-by-string word spotting are introduced: An end-to-end trainable model based on Faster R-CNN \cite{ren2017faster} and previous work \cite{wilkinson2015novel, wilkinson2016semantic}; and a simplified version that performs equally well or better in certain situations.
	\item Two novel data augmentation strategies for full manuscript pages, crucial for preventing model overfitting.
	\item Ablation studies that evaluate a set of model and training choices.
	\item An investigation into performance of the Region Proposal Network \cite{ren2017faster} as applied to manuscript images.
	\item State-of-the-art results on four datasets, including some very limited data settings.
	\item A case study conducted in collaboration with historians where we apply our model to a collection of 64 volumes of court records used for contemporary historical research. The result is an increase in processing speed and data size by orders of magnitude, compared to manual work.
\end{enumerate}

This paper is an extension of \cite{wilkinson2017neural}, where an initial version of this work was presented. Compared to the previous incarnation we have: introduced a simplified model (Section \ref{sec:model_mini}); carried out an ablation study (Section \ref{sec:ablation}); Improved upon previous results and extended experiments to two new datasets (Section \ref{sec:sota}); and extended the previous case study from 1 volume of court records to 64 (Section \ref{sec:case_study}), significantly increasing its scope, historical value, and complexity.

\section{Related Work}
The work in this paper is based on, and related to, a few different fields that we will briefly review in this section.

\subsection{Word spotting} \label{sec:ws_related_work}
Since its introduction over 20 years ago, word spotting \cite{WordSpotting1996} in manuscript images has come a long way. The initial approach uses template matching, with the image itself as a feature descriptor. Subsequent work introduced the idea of viewing the images as sequences of column features and applying sequence matching methods, in particular Dynamic Time Warping (DTW) \cite{kolcz2000line, RathManmantha2003, wahlberg2011data}, Hidden Markov Models (HMMs) \cite{rodriguez2009handwritten, fischer2012lexicon, rothacker2013bag}, and to a lesser extent, Recurrent Neural Networks (RNNs) \cite{frinken2012novel}. 



However, with the size of data ever increasing, the inefficiencies of sequence-based methods were becoming prohibitive. As a result, there was renewed interest in compact, fixed-length representations that allow for fast Euclidean distance calculations. \cite{rusinol2011browsing} build Bag-of-Visual-Words (BoVW) \cite{csurka2004visual} features on top of HOG descriptors, whereas \cite{almazan2014segmentation, LSA_embedding} use SIFT descriptors. In \cite{almazan2014word}, Fisher Vectors \cite{perronnin2007fisher} were built on top of SIFT descriptors.


Moreover, \cite{almazan2014word} and \cite{LSA_embedding} both allow for QbS word spotting using similar approaches. In \cite{LSA_embedding}, the authors create a textual descriptor based on character n-grams and use Latent Semantic Analysis \cite{deerwester1990indexing} to perform multi-modal fusion, mapping their visual BoVW and textual representations to the same space. In a similar vein, \cite{almazan2014word} use an attribute representation \cite{lampert2014attribute, farhadi2009describing} called Pyramidal Histogram of Characters (PHOC) as textual descriptor. They use Canonical Correlation Analysis (CCA) to learn a common subspace for the textual PHOC descriptor and visual Fisher Vectors. As it turns out, the approach in \cite{almazan2014word} proved to be the stronger system and the PHOC attribute representation has been widely adopted by the word spotting community, and has even been put to good use in lexicon-based text recognition \cite{poznanski2016cnn}.



Since then, many methods extending it have been proposed. In \cite{ghosh2015query, Ghosh_word_spotting}, the approach in \cite{almazan2014word} is extended to the segmentation-free setting. The method proposed in \cite{KrishnanDeepFeatureEmbedding} replaces the Fisher Vectors by a convolutional neural network (CNN) and the PHOC representation is learned using SVMs. Two methods were  simultaneously proposed where the two step Fisher Vector and CCA approach is replaced with end-to-end trainable CNNs \cite{sudholdtPhocnet, wilkinson2016semantic}. This strand of work has been consolidated and improved upon in \cite{sudholt2017attribute, krishnan2018word}. 


A majority of the proposed handwritten word spotting methods assume that the words or text lines have been segmented, or that this is easily achieved. It turns out that for many manuscripts, especially historical ones, this is not a valid assumption. To remedy this, an increasing number of segmentation-free\footnote{Note that segmentation-free has, in the earlier OCR literature, referred to not segmenting the word into characters, rather than manuscript into words.} word spotting methods have been proposed \cite{leydier2007text, rusinol2011browsing, rothacker2013bag, almazan2014segmentation, kovalchuk2014simple, Ghosh_word_spotting, rothacker2015segmentation} 

\begin{figure*}[t!]
	\begin{center}
		\includegraphics[width=0.99\linewidth]{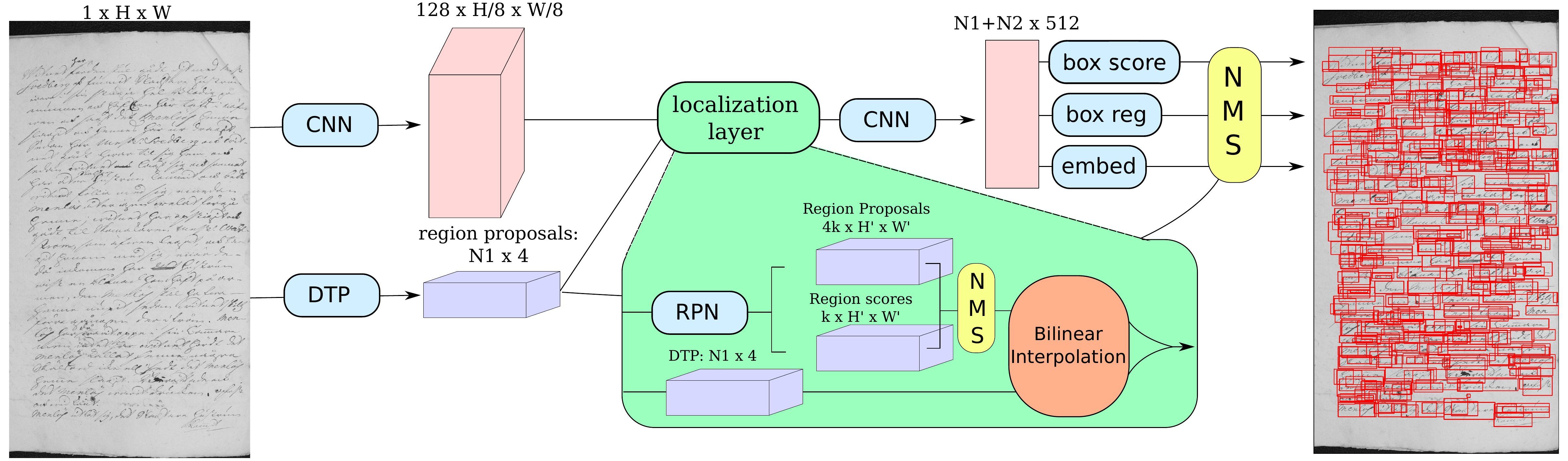}
	\end{center}
	\caption{The Ctrl-F-Net model at test time. Given an input image, it is fed through the first CNN of the model and Dilated Text Proposals (DTP) are extracted. These are then fed into the localization layer, where additional text proposals are extracted using a Region Proposal Network (RPN), followed by non-max suppression. The RPN proposals are added to the DTP proposals and fed through the Bilinear Interpolation layer, giving fixed length descriptors for each proposal. The proposals are fed through a second CNN and finally, each box coordinates are fine-tuned, given a wordness score, and a descriptor is extracted. Finally, a second non-max suppression is applied resulting a large number of region proposals, typically 2-3 times the number of ground truth boxes.}
	\label{fig:ctrlf_net}
\end{figure*}

Segmentation-free word spotting approaches can typically be placed into two broad categories based on how they generate regions from within an image. The first are methods based on a sliding window approach\cite{rusinol2011browsing, almazan2014segmentation, Ghosh_word_spotting, rothacker2015segmentation, rothacker2013bag}, where regions are generated positions along a regular grid over a manuscript page. The grid is usually either on the pixel level or on top of densely extracted features. This method is common in the QbE setting as the size of the query image in pixels known, allowing for constraints on the sizes of generated regions\cite{rusinol2011browsing,almazan2014segmentation, Ghosh_word_spotting,rothacker2013bag}. For QbS, the width and height of the region needs to be estimated, given the query\cite{rothacker2015segmentation}. The main drawback with these methods is the large amounts of regions generated, resulting in false positives and long processing times \cite{kovalchuk2014simple}. Additionally, due to a lack of attention, sliding window techniques are not robust against small shifts of the input, as this would lead to a misalignment av regions compared to ones extracted from the un-shifted input.

The second category are methods using connected components \cite{leydier2007text, ghosh2015query, kovalchuk2014simple, rothacker2017word, ghosh2018text}. In \cite{leydier2007text}, connected components in the shape of vertical strokes are extracted by performing mathematical morphology to separate characters. A popular approach is to binarize the image, extract connected components, and group them in a bottom-up fashion using heuristics and finally extract bounding boxes\cite{kovalchuk2014simple, ghosh2015query}. A similar approach is used in \cite{krishnan2016matching} for matching entire documents using distributions of word images. A combination of the two approaches is used in \cite{rothacker2017word}. Here, extremal regions are extracted on top of text presence scores computed using a sliding window. A related approach is applied to word segmentation in \cite{wilkinson2015novel}, where morphological closing is used using a variety of different kernel sizes to connect characters into words and extract bounding boxes. These methods still over-segment the manuscript, but the number of proposals is typically fewer than for sliding window based methods. They are also more robust to small input shifts as the connected components provide a kind of attention. The downsides include a sensitivity to physical degradations like ink blotches and difficulty with densely written manuscripts.

Two approaches for object detection were recently evaluated for the task of word segmentation in historical manuscripts \cite{moysset2018learning}. The first was YOLO \cite{redmon2016you} which did not yield any successful results, which was attributed to the assumption of only detecting one object per spatial cell. This is a reasonable assumption for natural images with several objects, but not for manuscript images with hundreds of words to detect. The second object detector was Multibox \cite{erhan2014scalable}, which managed to generate decent results, but was outperformed by the proposed method in \cite{moysset2018learning}. 

A recent approach that can be seen as hybrid between sliding window and connected components is presented in \cite{axler2018toward}. Here, the authors use a Resnet encoder-decoder network to produce a heatmap denoting each pixels probability of being inside, outside, or near a bounding box. The heatmap is smoothed using a smoother network before being fed to a Proposal Generation network that produces bounding box coordinates, which are subsequently filtered using a proposal Filter network.

\subsection{Attribute Representations and Label Embeddings}
Modern word spotting methods are closely related to two common tasks in computer vision. The widely adopted PHOC representation \cite{almazan2014word} is an attribute representation where each dimension corresponds to the presence or absence of a character in a part of a word. In the case of word spotting, the attributes are then used to retrieve words with similar attributes. Attribute representations have been successfully used in zero-shot learning \cite{lampert2014attribute, farhadi2009describing}

The recently introduced embedding for word spotting DCToW \cite{wilkinson2016semantic} is similar to PHOC in that it is hand-engineered, but it does not consist of binary attributes, but a low-frequency, real-valued representation of a text string. It shares a greater similarity with the Spatial Pyramid of Characters introduced for text recognition in \cite{rodriguez2015label}, which is similar to PHOC except that character occurrences are counted, not a binary presence/absence.

The attribute and label embedding representations allow seamless retrieval of words not present in the training data, known as zero-shot learning. This same technique is used for multi-label image classification \cite{chollet2016information}, and text recognition \cite{rodriguez2015label}. Moreover, word spotting shares similarities with approaches for multi-modal embeddings for zero-shot learning \cite{frome2013devise, socher2013zero}, with the difference that the text modality has a fixed embedding. 

\subsection{Scene Text Recognition}
The model proposed in this paper is similar to work in end-to-end scene text detection and recognition, which has been receiving increasing attention \cite{jaderberg2016reading, neumann2016real, li2017towards, busta2017deep, liu2018fots}. In \cite{jaderberg2016reading}, an end-to-end system for text localization, recognition and retrieval based on region proposals and CNNs is proposed. Another approach is made in \cite{neumann2016real}, where characters are detected and grouped together in a bottom-up fashion to build words and text lines. Two similar approaches \cite{busta2017deep, li2017towards} use region proposal networks (RPNs) to generate candidates for text regions and then transcribe them. The difference lies in  that \cite{busta2017deep} uses the Connectionist Temporal Classification (CTC) loss \cite{graves2006connectionist} to decode a region whereas \cite{li2017towards} employs a Recurrent Neural Network (RNN). Similarly, \cite{liu2018fots} makes us of an RPN and a CTC loss, but with a novel ROIRotate operation that maps arbitrarily oriented region proposals to axis-aligned feature maps. Our proposed models are similar to the RPN-based scene text recognition models, where the greatest difference lies in that we learn to embed word images in a word embedding space where they perform text recognition.

\section{The Models}
In this section we introduce the two models used in this paper. The first is the Ctrl-F-Net that we introduced in the previous version of this paper \cite{wilkinson2017neural}. The second model is a simplified version of Ctrl-F-Net we call Ctrl-F-Mini. In section \ref{sec:ablation}, we provide results from a series of ablation studies investigating the performance of the two models. 

\subsection{Ctrl-F-Net}
The model we propose is a deep convolutional neural network inspired by previous work on object detection\cite{ren2017faster}, dense image captioning \cite{densecap} and segmentation-based word spotting \cite{wilkinson2016semantic}. We call it Ctrl-F-Net, named after the well known shortcut for word search in many word processors. It is a model that simultaneously proposes and scores word candidate region proposals and embeds them into a word embedding space, wherein a search can be performed. The input to the model is a full manuscript page. The output is set of bounding box region proposals, their scores that correspond to the probability of containing a word, and an embedding. Optional external region proposals \cite{wilkinson2015novel} can be added as inputs and be used during both training and testing. It turns out that this increases the performs, see section \ref{sec:experiments}. A total of five loss functions are used, two in the middle of the model, and three towards the end, which lets the model learn all the tasks at hand. Figure \ref{fig:ctrlf_net} contains an overview of the model.

A grayscale input image of size $H, W$ first resized so that $max(H, W) = 1720$, while keeping the aspect ratio intact. Then it is fed through several layers of a CNN, until it has been spatially downsampled by a factor of 8. As the input is a full manuscript page, a 34-layer pre-activation ResNet \cite{he2016identity} is used as the CNN architecture due to its small memory footprint while still achieving high performance. The feature maps are then fed through a \emph{localization module}, that consists of: i) an RPN that generates region proposals and corresponding scores; ii) A non-max suppression (NMS) step to remove redundant, overlapping proposals; iii) A resizing layer produces fixed sized outputs given variable sized inputs; iv) loss functions for the region proposals coordinates and scores. 

\begin{figure*}[t!]
	\begin{center}
		\includegraphics[width=0.99\linewidth]{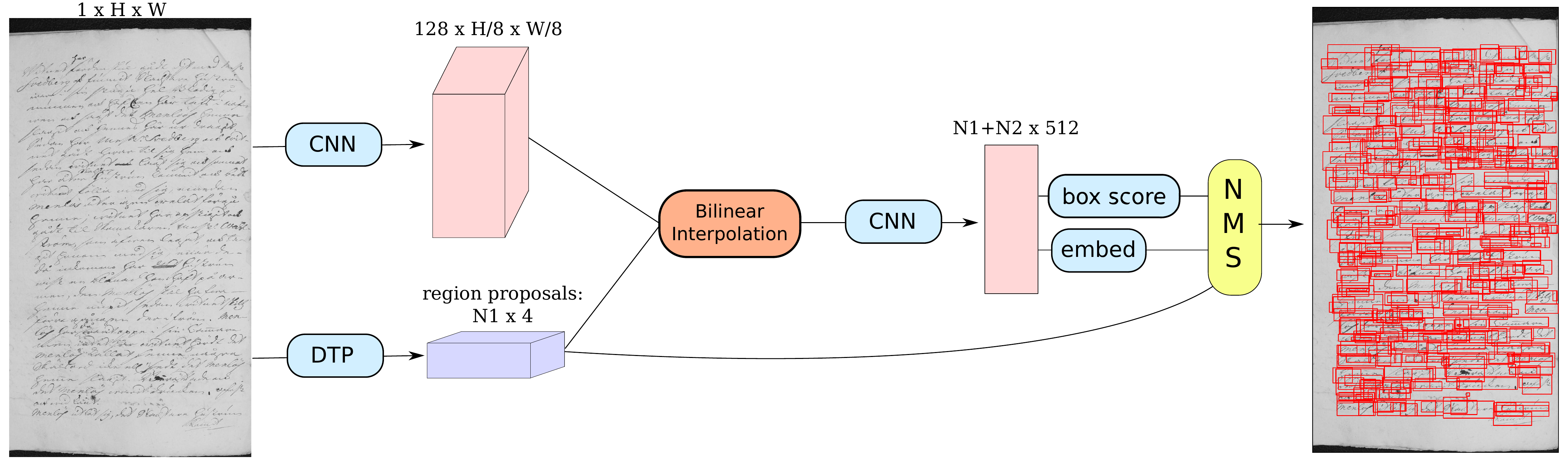}
	\end{center}
	\caption{Ctrl-F-Mini at test time. Given an input image, it is fed through the first CNN of the model and Dilated Text Proposals (DTP) are extracted. These are then fed through the bilinear interpolation layer, which resizes boxes to a fixed size. The proposals are fed through the rest of the CNN and a wordness score and a descriptor is extracted. Finally, non-max suppression is applied.}
	\label{fig:ctrlf_net_mini}
\end{figure*}

The input feature maps are first fed to an RPN, with one branch that regresses K=15 anchor boxes using a convolutional layer with $4K$ output channels. To accommodate the varying aspect ratios of words, the anchor boxes have the sizes $\{20, 40, 60\} \times \{ 30, 90, 150, 210, 300\}$, in pixels. A second parallel branch predicts \emph{wordness} scores for each box, that represent the probability that a box is situated atop a word. The boxes and scores are then combined and NMS is applied. The RPN can be seen to fall into the category of sliding window based segmentation-free word spotting methods. Proposals with an Intersection-over-union (IoU) overlap with a ground truth bounding box greater than 0.75 are considered as positives and  negatives are defined as proposals with an IoU lower than 0.4. Proposals that fall between are ignored. A total of 256 proposals, 128 positives and negatives, are sampled and are used to calculate the mid-network wordness and regression losses.

The boxes of varying size sampled from the RPN are then resized using Bilinear Interpolation \cite{jaderberg2015spatial, densecap} to a fixed output size of $8 \times 20$ pixels. They are then fed through the rest of the CNN and used as input to three parallel branches. The first branch is fully connected (FC) layer with 4 outputs that refines the box coordinates by regressing them once again. Similarly, the second branch is also a FC layer with a single output that predict the final wordness scores. The third branch is a small FC embedding network with 2 hidden layers, that does the final embedding. 

For the mid-network scores and output wordness scores, we use a binary logistic loss. The bounding boxes are parameterized according to \cite{girshick2015fast}, both for the anchor box regression and the output box regression. The boxes are represented as the quadruples $(x_c, y_c, w, h)$, where $x_c$ and $y_c$ are the center of a box and $w$ and $h$ is its width and height. The functions to learn are normalized translation offsets for $x$ and $y$ and log-space scaling factors for $w$ and $h$.
The loss is a smooth $l1$ loss, also known as a specialized version of the Huber loss \cite{huber1964robust}

\begin{equation}
L_{reg}(x_i, t_i) =
\left\{
\begin{array}{lcr}
\frac{1}{2} (x_i - t_i)^2  &\textnormal{if }  |x_i - t_i| & < 1 \\[5pt]

|x_i - t_i| - \frac{1}{2}  &\textnormal{if }  |x_i - t_i| & \geq 1 \\
\end{array}
\right.
\end{equation}
where $x_i$ is one of $\{x_c, y_c, w, h\}$ and $t_i$ is its corresponding target.

The embedding branch is a fully-connected network with two hidden layers of size 4096, with batch normalization\cite{ioffe2015batch} after each layer followed by the hyperbolic tangent activation function. The final layer is an $l^2$-normalization layer. It only receives the regions labelled positive as input. 

As loss function for the embedding network, we use the Cosine Embedding loss that has successfully been used in segmentation-based word spotting \cite{wilkinson2016semantic}. It is defined as

\begin{equation}\label{eq:cosine_embedding_loss}
L_{emb}(\mathbf{u}, \mathbf{v}, y) =
\left\{
\begin{array}{lcr}
1 - \frac{\mathbf{u}^\mathsf{T}\mathbf{v} }{||\mathbf{u}|| \cdot ||\mathbf{v}||} &\textnormal{if }  y = 1 \\[6pt]
\max(0, \frac{\mathbf{u}^\mathsf{T}\mathbf{v} }{||\mathbf{u}|| \cdot ||\mathbf{v}||} - \gamma) &\textnormal{if }  y = 0
\end{array}
\right.
\end{equation}

where $\mathbf{v}$ is an embedding of a positive region proposal and $\mathbf{u}$ is a ground truth embedding. If $y=1$, $\mathbf{v}$ and $\mathbf{u}$ match, and they are moved closer together. If $y=0$, they do not match and $\mathbf{v}$ and $\mathbf{u}$ are moved further apart. 
%
%
%

The total loss function is a weighted linear combination of the five losses. 
\begin{align}
\begin{split}
	L_{tot} = 10^{-2} &\cdot (L_{rpn\_reg} + L_{rpn\_score}) + \\[5pt] 
			  10^{-1} &\cdot (L_{reg} + L_{score}) + 3 \cdot L_{emb}
\end{split}
\end{align}

\subsection{Ctrl-F-Mini}\label{sec:model_mini}
In order to evaluate different model choices, in particular the source of region proposals, we introduce the Ctrl-F-Mini model. The main difference to the full model is the removal of the region proposal network, which in turn leads to other parts that are no longer relevant. From the localization module, only the bilinear interpolation remains. Towards the end, the end box scoring and embedding branches are kept. Ctrl-F-Mini is trained using external region proposals. The mid-network box scoring and regression losses and the end box regression loss is removed. The reduced model can be seen in Figure \ref{fig:ctrlf_net_mini}. This results in a greatly simplified model, which takes about a quarter of the time to train. Furthermore, inference becomes a lot faster as fewer region proposals are used, and the computations related to the RPN (in particular the non-max suppression over several hundred of thousand proposals) are not performed.

\subsection{Querying} \label{sec:querying}
During inference, a manuscript page and $N_1$ optional external region proposals are fed through the model that outputs: an $N \times 4$ matrix of region proposals; an $N \times D$ matrix of descriptors, where $D$ is the dimensionality of the word embedding that is used; and an $N\textnormal{-dimensional}$ vector of wordness scores, where for Ctrl-F-Net $N = N_1 + N_2$. We typically set $N_2 = N_1$ on a page-by-page basis ($N = N_1$ Ctrl-F-Net mini). We then threshold the wordness scores, only keeping proposals with a score $>t_s$, followed by an NMS step using an overlap threshold $t_{nms}$.

Once a query is selected, either by cropping a part of an image for QbE, or providing a search query for QbS, it is first transformed to the word embedding space. Then the cosine distance is used to compare the query to each region proposal and they are sorted w.r.t. their similarity to the query. Using the similarity the query as a score, we perform a final NMS step with an overlap threshold set to 0.

\subsection{Word Embeddings}
\begin{figure}[t!]
	\begin{center}
		\includegraphics[height=0.37\linewidth]{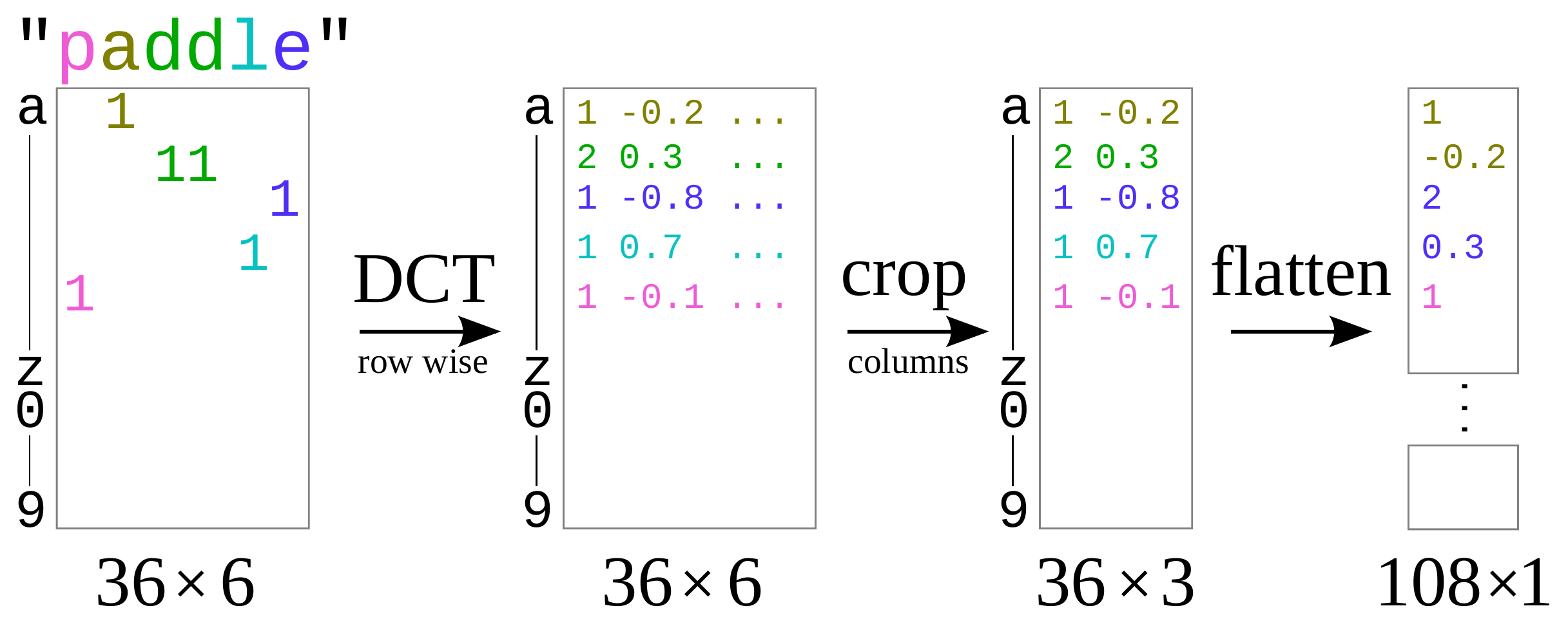} \\
		\vspace{0.5cm}
		\includegraphics[height=0.37\linewidth]{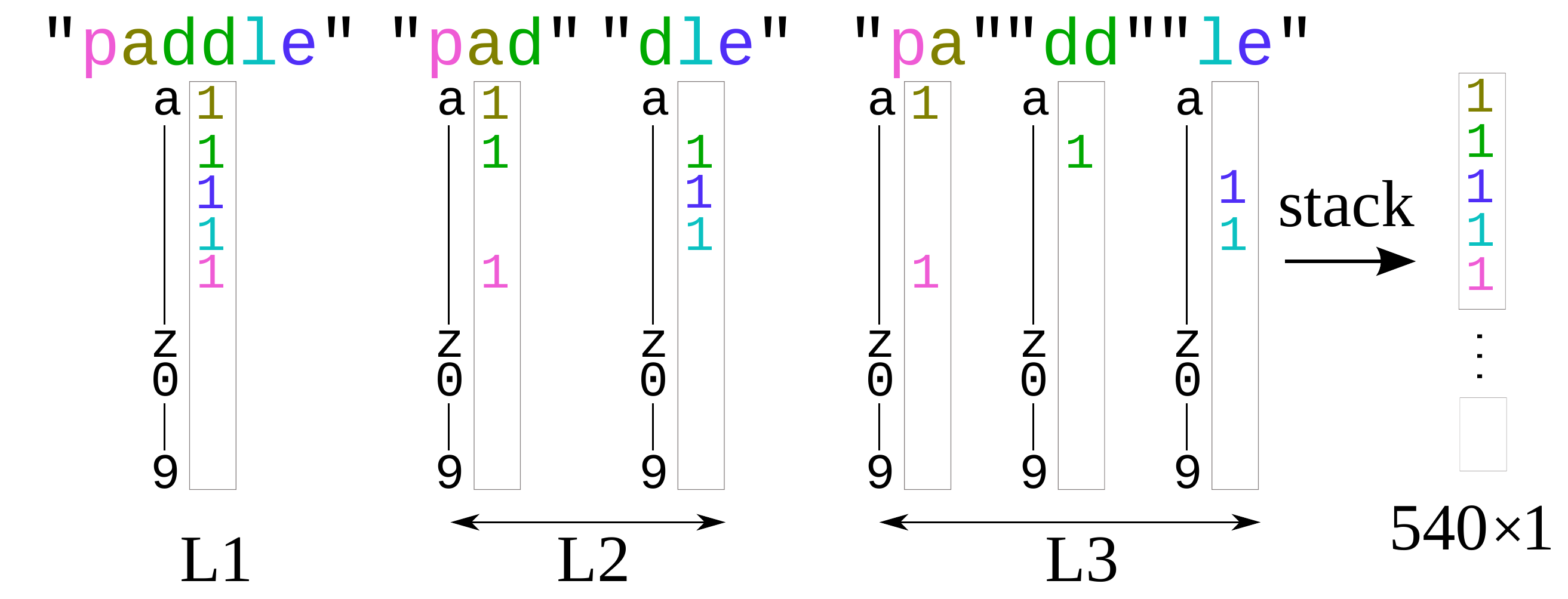}
	\end{center}
	\caption{The two word embeddings evaluated in this paper, DCToW (top) and PHOC (bottom). Note that we only show 3 of the 5 levels of the PHOC embedding here.}
	\label{fig:embeddings}
\end{figure}

In the recent word spotting literature based on using word embeddings, two have been most successful, see Figure \ref{fig:embeddings}. The first, and by far most popular, is the Pyramidal Histogram of Characters, or PHOC \cite{almazan2014word, krishnan2018word, sudholt2017attribute, ghosh2018text}. Provided a text string, the number of pyramid levels, and an alphabet of length $K=36$ (we use the digits 0-9 and lower-case letters for all experiments), construct a binary occurrence vector for each sub-word in each level of the pyramid (we use pyramid levels 1-5), and concatenate them. The resulting vector $\mathbf{u}_p$ is a $36 \cdot (1 + 2 + 3 + 4 + 5) = 540$ dimensional binary vector, i.e., $\mathbf{u}_p \in \{0, 1\}^{540}$. The earliest papers using PHOC augmented the alphabet with the most common bi-grams for a particular language. Subsequent work have achieved better results without them while keeping the embedding language agnostic, and we do the same.

The second embedding is the Discrete Cosine Transform of Words (DCToW), recently introduced in \cite{wilkinson2016semantic}. It is a low frequency, distributed representation of a word, that has recently achieved state-of-the-art results in segmentation-based word spotting. Given a word of length $m$ and an alphabet of length $K$, first build a $m \times K$ one-hot matrix representation. Then apply the Discrete Cosine Transform (DCT) to each row of the matrix. Finally, keep the first $r$ components of the DCT, and flatten the $m \times K$ matrix into an $r \cdot K$ dimensional vector, $\mathbf{u}_d$. Following \cite{wilkinson2016semantic}, we set $r=3$ making $\mathbf{u}_d \in {\rm I\!R}^{108}$. Words that are shorter than $r$ characters are padded with zeros to get the correct length.

\subsection{Dilated Text Proposals}
The ideal case for a segmentation-free word spotting system is maximizing the recall while keeping the number of proposals as low as possible. Referring back to the distinctions made in section \ref{sec:ws_related_work}, the RPN would fall into the category of sliding window approaches. As such, a likely improvement of the recall can be achieved by using a complementary external region proposal method based on connected components. We use the approach from \cite{wilkinson2015novel}, which we call Dilated Text Proposals (DTP) for sake of clarity. Given a grayscale image, DTP first creates a set of $j$ binary image by thresholding at $j$ different multiples of the image mean value. Then applies morphological closing to each binary image using a set of $l$ generated rectangular kernels. For each of the $j \cdot l$ images, find the connected components, then extract bounding boxes for each connected component and remove duplicate boxes.

\section{Data Augmentation}
\begin{figure}[t!]
	\begin{center}
		\includegraphics[width=0.45\linewidth]{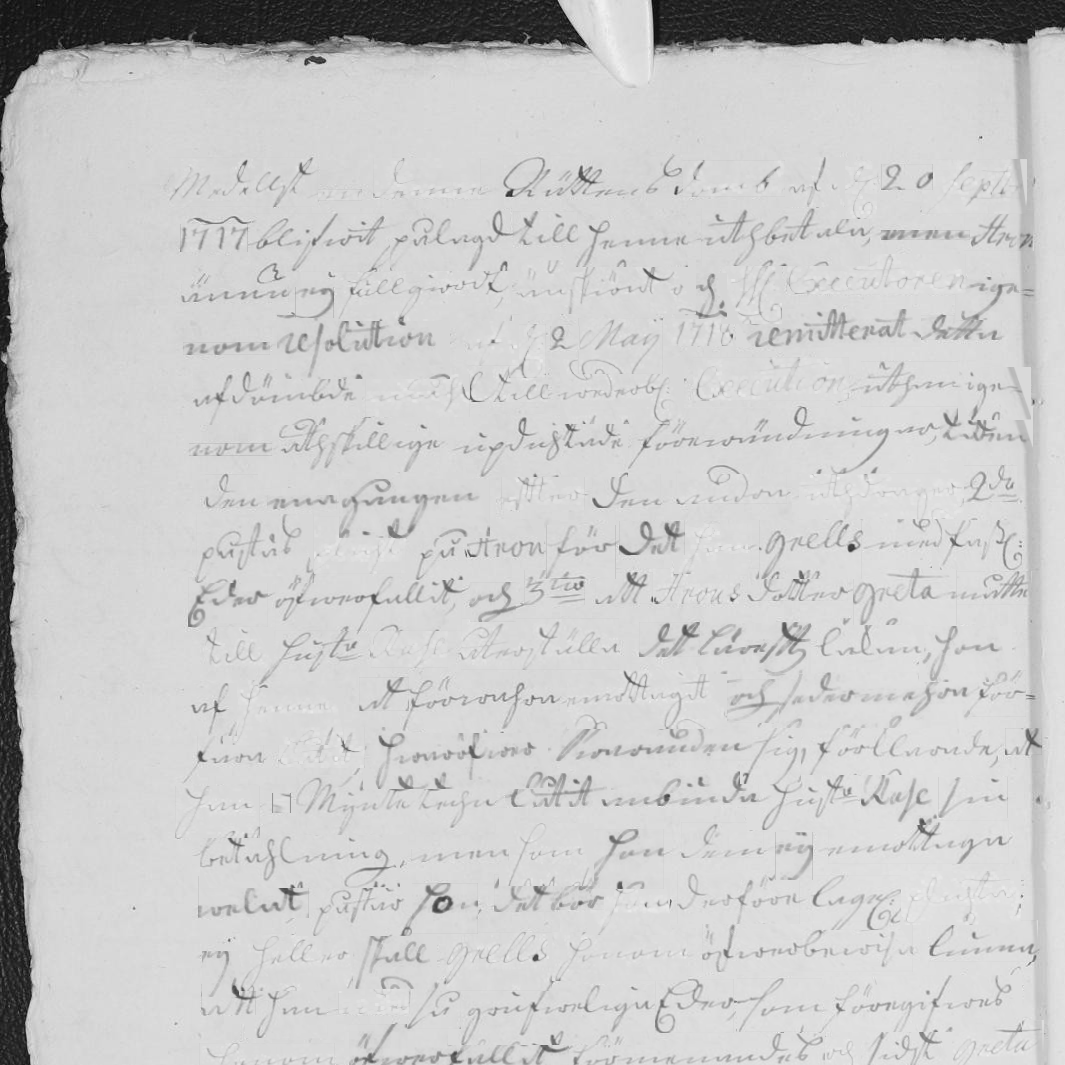}
		\hspace{0.2cm}
		\includegraphics[width=0.45\linewidth]{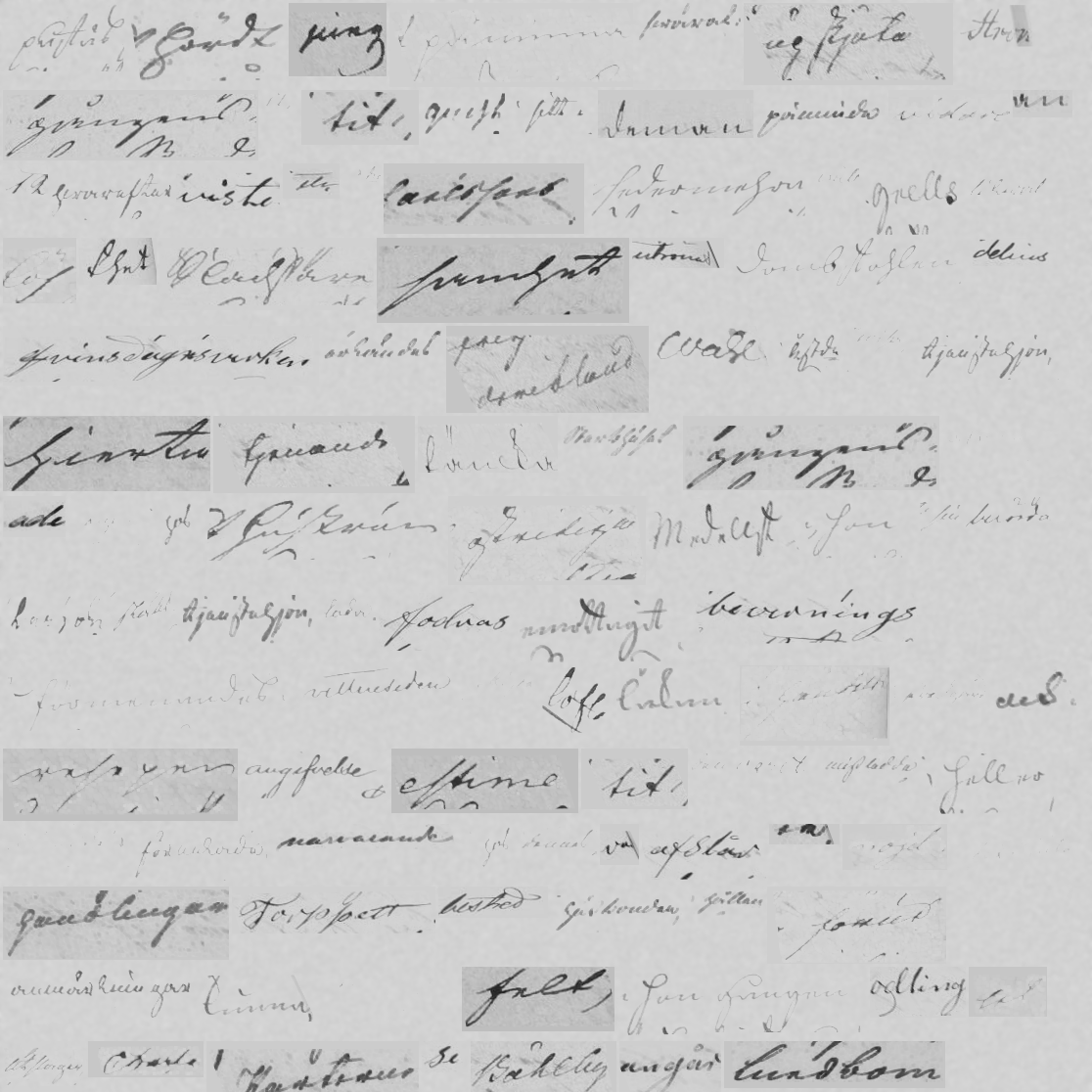}
	\end{center}
	\caption{A visual comparison between in-place (left) and full page augmentation (right). In-place augmentation provides style variation while full page augmentation allows us to control word class distributions.}
	\label{fig:data_aug}
\end{figure}

As we are operating in a small data setting (as few as 5 manuscript pages for training), data augmentation is crucial to prevent severe overfitting on the training data. We propose two complementary ways of augmenting the entire manuscript pages that we call \emph{full-page} and \emph{in-place} augmentation. The two techniques are visually compared in Figure \ref{fig:data_aug}.

Full-page augmentation allows to have control over the distribution of classes, which is important for learning a discriminative word embedding. It works by uniformly sampling word images from the training set, augmenting them, and placing the row-by-row on background canvas. We adopt the affine and grayscale morphology augmentation from \cite{wilkinson2016semantic}. The canvas is created by uniformly sampling a background colour from an interval centered on the median of all images in the training set and adding on some Gaussian noise. The finished augmented page looks like left-aligned manuscripts of randomly sampled word images.

In-place augmentation is designed to keep the overall look of the page intact, while still providing some useful variation in writing style. Ideally, this helps the model generate and score region proposals, while still providing variation for the learning word embeddings, although without control of class distributions. For a given manuscript page, we iterate through the ground truth bounding boxes and augment each word image in-place. We apply a shearing transform and followed by grayscale morphological dilation or erosion while ensuring the output has the same size as the input so that it can be slotted back into place.

\section{Experiments}\label{sec:experiments}
Here, we perform the main quantitative evaluation of our models, including an ablation study, on two widely used benchmarks for word spotting.

\textbf{The George Washington (GW) Dataset \cite{lavrenko2004holistic}} was written in English the middle of the 18\textsuperscript{th} century by George Washington and his secretaries. It consists of 20 pages, or 4860 words. We follow the evaluation procedure used in \cite{rothacker2015segmentation}, by using two different splits of the pages into training, validation and test sets. For the first split called GW 15-5, has a training set of 15 pages and 5 for testing. The second is a 5-15 split with 5 training and 15 test pages. In both cases, we use 1 page as a validation set. The reported results are the average of four cross validations. The bounding boxes for the GW are manually annoatated, resulting in a large amount of extra space around 1-2 character words, causing a significant decrease in recall for higher overlap thresholds. To counteract this annotation issue, we pad the DTP proposals with 10 pixels for this dataset.

\textbf{The IAM Offline Handwriting Dataset \cite{marti2002iam}} is a modern cursive dataset consisting of 1539 pages, or 115320 words, written by 657 writers. We use the official train/val/test split for \emph{writer independent text line recognition}, where there is no writer overlap between the different splits. Following standard protocol, we remove stop words from the set of queries, and in line with \cite{almazan2014word}, queries that come from lines that are marked as containing segmentation errors are removed. Ground truth boxes that are so small that they collapse to a width or height of zero when downsampled by a factor 8 are also removed.

\textbf{The Botany and Konzilsprotokolle Datasets} were introduced in the ICFHR 2016 Handwritten Keyword Spotting competition \cite{pratikakis2016icfhr2016}. The collections are from the 19\textsuperscript{th} and 17\textsuperscript{th} centuries respectively. Following \cite{rothacker2017word}, we use the official \texttt{Train III} set for training. It contains 114 and 45 images for each respective dataset. We select 5 and 1 images respectively as validation set for the datasets. The test set for both datasets are 20 pages. We use the set of queries defined for the competition as well as the official software to calculate results.

\begin{table*}[t!]
	\renewcommand{\arraystretch}{1.3}
	\caption{Ablation results for different model variants the GW 15-5 and IAM datasets. Recall is calculated based on the proposals left after the final NMS stage.}
	\label{tab:ablation}
	\centering
	\begin{tabular}{llcccccccccccc}
		
		& & \multicolumn{6}{c}{GW 15-5}  &  \multicolumn{6}{c}{IAM} \\\cmidrule(r){3-8} 
		\cmidrule(l){9-14} 
		& & \multicolumn{2}{c}{MAP 50\%} & \multicolumn{2}{c}{MAP 25\%} & \multicolumn{2}{c}{Recall} & \multicolumn{2}{c}{MAP 50\%} & \multicolumn{2}{c}{MAP 25\%}& \multicolumn{2}{c}{Recall} \\ \cmidrule(r){3-4}\cmidrule(lr){5-6}\cmidrule(lr){7-8}\cmidrule(lr){9-10}\cmidrule(lr){11-12}\cmidrule(l){13-14}
		Model Variant & Embedding & QbE & QbS & QbE & QbS & 50\% & 25\% & QbE & QbS& QbE & QbS & 50\% & 25\% \\
		\thickhline
		\addlinespace[1.5pt]
		\multicolumn{14}{c}{Baselines from \cite{wilkinson2017neural}}\\
		\addlinespace[1.5pt]
		\hline
		Ctrl-F-Net & DCToW & 90.5 & 91.0 & 97.0 & 95.2 & \textbf{99.4} & 99.9 & 72.0 & 80.3 & 74.1 & 82.5 & 98.1 & 98.9 \\
		Ctrl-F-Net & PHOC & 90.9 & 90.1 & 96.7 & 93.9 & - & - & 71.5 & 78.8 & 73.7 & 80.8 & - & - \\
		\hline
		\addlinespace[1.5pt]
		\multicolumn{14}{c}{Baselines}\\
		\addlinespace[1.5pt]
		\hline
		Ctrl-F-Net & DCToW & 91.9 & 92.9 & 96.8 & 95.7 & 93.8 & 99.5 & 73.9 & 81.9 & 77.2 & 85.3 & 97.6 & 98.5\\
		Ctrl-F-Mini & DCToW & \textbf{92.5} & 93.4 & 96.9 & 96.2 & 95.6 & 99.0 & 73.9 & 83.2 & 75.8 & 85.1 & 93.2 & 94.3\\
		Ctrl-F-Net & PHOC & 90.8 & 90.5 & 97.1 & 95.3 & 96.7 & 99.8 & 74.6 & 83.0 & 78.6 & 87.4 & 98.0 & 98.9 \\
		Ctrl-F-Mini & PHOC & 92.2 & 92.9 & 97.1 & 96.1 & 96.1 & 99.2 & 74.6 & 84.9 & 77.0 & 87.0 & 93.1 & 94.2 \\
		\hline
		\addlinespace[1.5pt]
		\multicolumn{14}{c}{Cosine loss}\\
		\addlinespace[1.5pt]
		\hline
		Ctrl-F-Net&  DCToW & 91.8 & 92.3 & 96.9 & 95.4 & 95.8 & 99.6 & 74.7 & 83.7 & 77.7 & 86.7 & 97.2 & 98.2 \\
		Ctrl-F-Mini & DCToW & \textbf{92.5} & \textbf{93.5} & 97.2 & 96.3 & 96.1 & 99.0 & 74.9 & 84.9 & 77.2 & 86.8 & 92.5 & 93.7 \\
		Ctrl-F-Net & PHOC & 91.7 & 92.3 & 96.6 & 95.1 & 95.1 & 99.4 & 74.0 & 83.0 & 78.3 & 86.8 & 98.5 & 99.3 \\
		Ctrl-F-Mini & PHOC & 91.6 & 91.8 & 97.0 & 96.2 & 96.5 & 99.2 & 75.7 & \textbf{86.1} & 77.8 & 87.9 & 91.7 & 92.9 \\
		\hline
		\addlinespace[1.5pt]
		\multicolumn{14}{c}{Use DTP proposals during training}\\
		\addlinespace[1.5pt]
		\hline
		Ctrl-F-Net & DCToW & 89.7 & 89.2 & \textbf{97.6} & \textbf{96.8} & 98.2 & 99.8 & 73.4 & 83.4 & 76.4 & 85.8 & 92.0 & 94.2 \\
		Ctrl-F-Net & PHOC & 91.4 & 93.2 & 96.9 & 96.0 & 94.5 & 99.3 & 72.4 & 82.0 & 78.3 & \textbf{88.1} & 92.3 & 93.6\\
		\hline
		\addlinespace[1.5pt]
		\multicolumn{14}{c}{RPN proposals only}\\
		\addlinespace[1.5pt]
		\hline
		Ctrl-F-Net & DCToW & 80.5 & 79.4 & 94.0 & 90.6 & 96.5 & 99.6 & 49.7 & 56.3 & 61.3 & 67.7 & 64.39 & 79.10 \\
		Ctrl-F-Net & PHOC & 80.8 & 79.7 & 94.1 & 90.8 & 97.2 & \textbf{100.0} & 53.8 & 62.7 & 69.6 & 77.3 & 64.04 & 81.90 \\
		\hline
		\addlinespace[1.5pt]
		\multicolumn{14}{c}{Binary Cross Entropy loss}\\
		\addlinespace[1.5pt]
		\hline
		Ctrl-F-Net & PHOC & 90.9 & 86.2 & 95.7 & 88.3 & 94.8 & 99.3 & \textbf{79.1} & 72.1 & \textbf{81.5} & 74.5 & \textbf{98.7} & \textbf{99.4} \\
		Ctrl-F-Mini & PHOC & 91.8 & 86.6 & 96.2 & 88.9 & 94.9 & 98.5 & 73.3 & 69.5 & 75.1 & 70.8 & 92.8 & 93.9 \\
		\thickhline
	\end{tabular}
\end{table*}

\subsection{Training} \label{sec:training}
The models are trained in a single phase. We first train a model (with weights initialized randomly using \cite{he2015delving} for convolutional layers, and a zero-mean Gaussian with a standard deviation of 0.01 for fully connected layers) using the synthetic IIIT-HWS-10k dataset \cite{krishnan2016matching}. Since it only consists of word images, we use the full-page augmentation technique to create 3000 synthetic document images. This model was used to initialize all other models.

For the other datasets, we create 5000 augmented images, split evenly between in-place and full-page augmentation, and add them to the original data. The input image is rescaled such that its longest side is 1720. We train each model for a maximum of 25000 iterations, and the measure the performance on a held out validation set every 1000 iterations. The model with the highest validation MAP score is used for testing. The learning rate is initially set to $10^{-3}$ for all models (except for Ctrl-F-Net on IAM which starts at $2\cdot10^{-4}$) and is multiplied every 10000 iterations by 0.1. We use ADAM \cite{kingma2014adam} to update the weights. Our implementation \footnote{\url{https://github.com/tomfalainen/neural-word-search}} is in Pytorch \cite{paszke2017automatic} and training time is approximately 10 hours for Ctrl-F-Net and 3 hours for Ctrl-F-Mini on an NVIDIA Titan GTX.

\subsection{Evaluation}
For the GW and IAM datasets, we evaluate our model using the standard metric used for word spotting, Mean Average Precision (MAP), where the Average Precision for a collection of size $N$ is defined as
\begin{equation}
AP = \frac{\sum_{k=1}^{N} P_k \cdot r_k}{R}
\end{equation}
where $P(k)$ is the precision measured at cut-off $k$ in the returned list, $R$ is the number of relevant results, and $r(k)$ is an indicator function that is 1 if a returned result at rank $k$ is relevant, and 0 otherwise. A retrieved word is considered relevant if its IoU overlap with a ground truth box is greater than a threshold $t_o \in \{0.25, 0.5\}$ and the label matches the query. The MAP score is the mean of the AP over the set of queries
\begin{equation}
MAP = \frac{\sum_{q=1}^{Q} AP(q)}{Q}
\end{equation}
where $Q$ is the number of queries. Unless stated otherwise, for QbE evaluation all the ground truth segmented word images in the test set is used. For QbS, all unique ground truth labels are used. Some methods, in particular \cite{ghosh2015query, Ghosh_word_spotting, ghosh2018text}, use a slightly different protocol for the GW dataset. Here all word instances in the dataset are used as queries for QbE, and all unique labels for QbS. The search is performed in all 20 pages. We perform a grid search over the score NMS overlap threshold $t_{nms}$, score threshold, and RPN score NMS overlap threshold when applicable. 

\subsection{Ablation and analysis}\label{sec:ablation}
To perform the ablation study, we adopt the methodology of using baseline model settings and changing one setting at a time, and always comparing with the baseline. We investigate the quantitative performance of various model choices, with the most significant being the source of region proposals (RPN, DTP, or both). Other experiments include comparing the PHOC and DCToW and three embedding loss functions. Recent work \cite{sudholt2017evaluating} argues that the Cosine loss \cite{chollet2016information} has outperformed other common loss functions for segmentation-based word spotting, including the Cosine Embedding loss. We evaluate this loss in the segmentation-free setting. The Cosine loss is defined as 

\begin{equation}
L(\mathbf{u}, \mathbf{v}) = 1 -  \frac{\mathbf{u}^\mathsf{T}\mathbf{v} }{||\mathbf{u}|| \cdot ||\mathbf{v}||}
\end{equation}

where $\mathbf{v}$ is an embedding of a positive region and $\mathbf{u}$ is the embedding of the ground truth label. It is the part of the Cosine Embedding loss (Equation \ref{eq:cosine_embedding_loss}) where $y=1$ . We also evaluate the Binary Cross Entropy (BCE) loss, which models the embedding as a multi-label binary classification problem, and is a common choice of loss function for the PHOC embedding. As the BCE loss requires a binary embedding, it is not applicable to the DCToW. We use the GW 15-5 and IAM datasets to evaluating the different model choices. 

%
%
%

\begin{table}
	\renewcommand{\arraystretch}{1.3}
	\caption{Recall comparison in \%, averaged over pages between the region proposal network and dilated text proposals using Ctrl-F-Net with the DCToW embedding. Filtered refers to the score thresholding and non-max suppression steps described in section \ref{sec:querying}.}
	\label{tab:recall}
	\centering
	\begin{tabular}{lccccccc}
		
		&&\multicolumn{2}{c}{GW 15-5} & \multicolumn{2}{c}{GW 5-15} & \multicolumn{2}{c}{IAM} \\ \cmidrule(lr){3-4}  \cmidrule(lr){5-6} \cmidrule(lr){7-8} 
		Method & Filtered & 50\%& 25\%& 50\%& 25\%& 50\%& 25\% \\
		\thickhline
		RPN     & & 98.3 & 100.0 & 98.6 & 99.9 & 69.1 & 86.5 \\
		DTP     & & 98.8 & 99.9 & 98.8 & 99.9 & 98.6 & 99.4 \\
		Combined& & 99.9 & 100.0 & 99.9 & 100.0 & 99.0 & 99.7 \\
		
		RPN     & \checkmark & 4.2 & 6.1 & 3.1 & 8.7 & 46.9 & 62.8 \\
		DTP     & \checkmark & 89.6 & 94.4 & 91.8 & 96.7 & 97.0 & 98.1 \\
		Combined& \checkmark & 93.8 & 99.5 & 94.9 & 99.7 & 97.6 & 98.5 \\
		\thickhline
	\end{tabular}
\end{table}

The top section of Table \ref{tab:ablation} contains the results from \cite{wilkinson2017neural}, which are a bit different from the new baselines. Since \cite{wilkinson2017neural} was published, we discovered a few mostly small bugs in our code, the most notable one is that the margin $\gamma$ for the Cosine Embedding loss was actually 0.2 instead of the reported 0.1, and the learning rate was accidentally multiplied by 0.1 in the first iteration of training, making the initial learning rate $2^{-4}$ instead of $2^{-3}$. 

The first variant we evaluated is using the Cosine loss instead of the Cosine Embedding loss. Although the results are not unanimous, the trend suggests that the Cosine loss is superior. This corroborates the findings in \cite{sudholt2017evaluating} that the Cosine loss is better for word spotting, and coupled with the fact that it is simpler makes the Cosine loss a better choice compared to the Cosine Embedding loss. From our experiments, the Binary Cross Entropy loss works very well on the QbE setting, notably getting the highest MAP score on IAM by a good margin, around 3\% for 50\% IoU. However, it underperforms when it comes to QbS.

The second set of experiments in Table \ref{tab:ablation} involve using both proposals from the RPN and DTP during training the Ctrl-F-Net. This is implemented as sampling both positive and negative boxes from the proposal pool from the DTP and RPN separately, and concatenating them before continuing the forward pass. Towards the end of the model, both sets of proposals are used for the box scoring and embedding losses, but only the RPN proposals are used for the end box regression loss. Although the results are mixed, this modification seems to give an improvement for the MAP using 25\% IoU.

To investigate the quality of the proposals w.r.t. MAP score, we only use RPN proposals to retrieve words from the manuscripts. There is a noticeable drop in performance for both GW and IAM datasets with this setup. Comparing proposal quality, the DTP only model i.e., Ctrl-F-Mini, clearly outperforms the RPN only model. Similarly, adding DTP proposals to the Ctrl-F-Net increases the MAP score by a good amount.

Across all experiments, the DCToW and the PHOC seem to work best on the GW and IAM datasets respectively. This suggests that it is best to evaluate both embeddings for the task at hand. All else being equal, the DCToW is preferable due to its smaller dimensionality, approximately a fifth of the PHOC. Finally, analysing the Ctrl-F-Net and Ctrl-F-Mini, the latter performs at a similar if not higher level compared to the former over all experiments. This comes at the cost of a slightly lower rate of recall, most notably on the IAM dataset.

The small difference in recall between the Ctrl-F-Net (both RPN and DTP) and Ctrl-F-Mini (only DTP) models merits further investigation into the contributions of the RPN and DTP proposals in the recall of Ctrl-F-Net. Table \ref{tab:recall} shows the recall rates of the two sources of region proposals, the RPN and the DTP, and their union using the baseline DCToW model on 15-5 and 5-15 GW datasets and the IAM dataset. The recall is the average over pages, and the number of proposals for each method is held the same. We evaluate the recall before and after filtering, which refers to the score thresholding and non-max suppression from Section \ref{sec:querying}. Before the filtering step, the recall is practically equal on the GW dataset. For the IAM dataset on the other hand, there is a noticeable gap between the two sources of region proposals. Post filtering, the gap widened considerably for the IAM dataset and a chasm appeared between the RPN and DTP on the GW dataset. Considering that it is the model that is scoring the proposals from the two sources, and the model was only trained on RPN proposals, for the model to favour the DTP heavily suggests that the DTP generates far superior proposals. This is further corroborated by the MAP results in Table \ref{tab:ablation}, where there is a significant drop in score and that the Ctrl-F-Mini is performing on a similar level to Ctrl-F-Net.

\begin{table}
	\renewcommand{\arraystretch}{1.3}
	\caption{MAP and recall comparison in \% on the Uppsala petitions dataset}
	\label{tab:petitions_dataset}
	\centering
	\begin{tabular}{lcccccc}
		&\multicolumn{6}{c}{Uppsala Petitions} \\ \cmidrule(lr){2-7}
		& \multicolumn{2}{c}{MAP 50\%} & \multicolumn{2}{c}{MAP 25\%} & \multicolumn{2}{c}{Recall} \\ \cmidrule(r){2-3}\cmidrule(lr){4-5}\cmidrule(lr){6-7}
		Method & QbE & QbS & QbE & QbS & 50\% & 25\% \\
		\thickhline
		Ctrl-F-Net & 47.4 & 33.0 & 56.8 & 37.8 & 86.9 & 93.4 \\
		Ctrl-F-Mini & 41.1 & 31.6 & 48.0 & 35.1 & 71.3 & 79.8 \\
		\thickhline
	\end{tabular}
\end{table}

\subsubsection{When to use the Ctrl-F-Net}
From the results of the ablation study, one may wonder why to use the Ctrl-F-Net at all when the Ctrl-F-Mini performs equally well but is faster and simpler. This question can also be formulated as: When to include the region proposal network? The key issue is how densely written, and therefore, how easily segmented the words are for a given manuscript collection. 

All the public benchmarks used for evaluation in this paper are relatively easily segmented. In fact, this is an issue with all widely used public word spotting benchmarks. This makes it difficult to evaluate the performance in terms of segmentation in a fair way. They therefore fail to encompass the important scenario of densely written, messy, crossed out text, which occur frequently in historical manuscripts. It is in this context the Ctrl-F-Net with its region proposal network can give a significant boost in recall over the Ctrl-F-Mini. The early modern court records in Section \ref{sec:case_study} contain many such pages. 

The DTP underperforms in terms of recall for these difficult pages because it has great trouble with separating intersecting text. We would argue that this is an issue for many (if not all) region proposal methods that are based on connected components (or detecting the ink on the page), such as the MSER based approach in \cite{rothacker2017word} or the approach from \cite{ghosh2018text} where they extract connected components from a thresholded image. The RPN has no such issues because it uses a sliding window, which is the main motivation for including it in the Ctrl-F-Net\footnote{Note that this is only just one way of mitigating the problem of these difficult-to-segment pages. There are undoubtedly other ways to tackle this issue.}. 

We conducted an additional experiment on a soon-to-be-released dataset of historical handwriting, which consists of source material from similar time period and geographical area, and written in the same style of writing. The main difference lies in their purpose, the dataset is made up of petitions to regional and national government in the form of letters. The dataset consists of 45 pages for training and 28 pages for testing. We train a Ctrl-F-Net and Ctrl-F-Mini (initialised with the model trained on the IIIT-HWS dataset), and evaluate their respective performance in terms of recall and MAP in Table \ref{tab:petitions_dataset}. We observe that in terms of MAP, the Ctrl-F-Net slightly but consistently outperforms the Ctrl-F-Mini. The same trend is true for recall, though the gap is much more pronounced.

In light of these results, we would recommend to use the Ctrl-F-Net (i.e., add the RPN) whenever you are working large and heterogeneous manuscript collections. For the public benchmarks, the largest discrepancies between the Ctrl-F-Net and Ctrl-F-Mini lie at the 50\% overlap threshold, however for the application we show in Section \ref{sec:case_study} where computation time is not crucial and results are manually reviewed by a user, the performance at 25\% overlap matters more. Here there is no discernible difference in MAP between the Ctrl-F-Net and Ctrl-F-Mini. 

\begin{table}
	\renewcommand{\arraystretch}{1.3}
	\caption{Inference (inf) and search times averaged over pages and queries respectively for each dataset in seconds and storage space requirements in megabytes.}
	\label{tab:query_time}
	\centering
	\begin{tabular}{lcccccc}
		&\multicolumn{3}{c}{Ctrl-F-Net} & \multicolumn{3}{c}{Ctrl-F-Mini}\\ \cmidrule(lr){2-4} \cmidrule(lr){5-7}
		Method & inf & search & space & inf & search & space \\
		\thickhline
		GW 15-5 & 12.45 & 0.06 & 0.96 & 3.69 & 0.12 & 1.4 \\
		GW 5-15 & 12.46 & 0.44 & 2.7 & 3.67 & 0.79 & 5.3 \\
		IAM & 1.75 & 5.09 & 24 & 0.50 & 7.69 & 19 \\
		Konz & 15.69 & 1.11 & 8.1 & 7.39 & 2.27 & 13 \\
		Botany & 11.75 & 3.53 & 16 & 3.42 & 3.23 & 12 \\
		\thickhline
	\end{tabular}
\end{table}

\begin{table*}[t!]
	\renewcommand{\arraystretch}{1.3}
	\caption{MAP comparison in \% with state-of-the-art segmentation-free methods on the GW and IAM datasets. The Ctrl-F-Net results marked with an asterisk use the evaluation protocol from \cite{Ghosh_word_spotting,ghosh2015query, ghosh2018text} (only relevant for GW 15-5).
	}
	\label{tab:sota}
	\centering
	\begin{tabular}{llcccccccccccc}
		
		& & \multicolumn{4}{c}{GW 15-5}  &  \multicolumn{4}{c}{GW 5-15} &  \multicolumn{4}{c}{IAM}\\\cmidrule(r){3-6} 
		\cmidrule(l){7-10} 	\cmidrule(l){11-14} 
		& & \multicolumn{2}{c}{MAP 50\%} & \multicolumn{2}{c}{MAP 25\%} & \multicolumn{2}{c}{MAP 50\%} & \multicolumn{2}{c}{MAP 25\%} & \multicolumn{2}{c}{MAP 50\%}& \multicolumn{2}{c}{MAP 25\%} \\ \cmidrule(r){3-4}\cmidrule(lr){5-6}\cmidrule(lr){7-8}\cmidrule(lr){9-10}\cmidrule(lr){11-12}\cmidrule(l){13-14}
		Method & Embedding & QbE & QbS & QbE & QbS & QbE & QbS & QbE & QbS& QbE & QbS & QbE & QbS \\
		\thickhline
		Ctrl-F-Net & DCToW & 90.9 & 91.7 & 97.0 & 96.2 & 84.1 & 76.2 & 94.0 & 83.8 & 75.4 & 85.5 & 77.4 & 87.0 \\
		Ctrl-F-Net & PHOC & 90.7 & 92.0 & \textbf{97.3} & \textbf{96.4} & 87.5 & 79.9 & 93.4 & 82.4 & 72.1 & 81.9 & 78.0 & \textbf{88.4} \\
		Ctrl-F-Mini & DCToW & \textbf{92.5} & \textbf{93.5} & 97.2 & 96.3 & 87.6 & 80.8 & \textbf{94.7} & 86.0 & 74.9 & 84.9 & 77.2 & 86.8 \\
		Ctrl-F-Mini & PHOC & 91.6 & 91.8 & 97.0 & 96.2 & \textbf{87.9} & \textbf{81.4} & 94.3 & \textbf{86.1} & 75.7 & \textbf{86.1} & 77.8 & 87.9 \\
		
		BoF HMMs \cite{rothacker2015segmentation} & n/a & - & 76.5 & - & 80.1 & - & 54.6 & - & 58.1 & - & - & - & -\\
		$\textnormal{AAM+SIFT}_{\textnormal{quant}}$ \cite{rothacker2017word} & PHOC & 81.6 & 84.6 & 92.0 & 90.6 & - & - & - & - & - & - & - & -\\
		Encoder-Decoder Net \cite{axler2018toward} & PHOC & - & - & - & - & - & - & - & - & - & 85.4 & - & 85.6\\
		Hwnet v2 Ctrl-F-Mini \cite{krishnan2019hwnet} & Hwnet & 92.0 & - & 96.7 & - & - & - & - & - & \textbf{82.0} & - & \textbf{82.4} & -\\
		\thickhline
		\addlinespace[1.5pt]
		\multicolumn{14}{c}{Alternate Evaluation Protocol}\\
		\addlinespace[1.5pt]
		\thickhline
		\addlinespace[2pt]
		Ctrl-F-Net* & DCToW & \textbf{83.1} &\textbf{ 84.7} & \textbf{97.1} & \textbf{94.5} &  - & - & - & - & - & - & - & -\\
		SW \cite{Ghosh_word_spotting}& PHOC & 67.7 & - & - & - & - & - & - & - & 42.1 & - & - & -\\
		BG index \cite{ghosh2015query} & PHOC & - & 73.3 & - & - & - & - & - & - & - & 48.6 & - & -\\
		SVM Fisher Vectors \cite{ghosh2018text} & PHOC & 77.2 & 69.9 & - & - & - & - & - & - & 38.7 & 44.7 & - & -\\
		\thickhline
	\end{tabular}
\end{table*}

\begin{table*}[h]
	\renewcommand{\arraystretch}{1.3}
	\caption{MAP comparison in \% with state-of-the-art segmentation-free methods on the Botany and Konzilsprotokolle datasets. Results marked with asterisk uses a smaller training data split.}
	\label{tab:sota_konz_botany}
	\centering
	\begin{tabular}{llcccccccc}
		
		& & \multicolumn{4}{c}{Botany}  &  \multicolumn{4}{c}{Konzilsprotokolle}  \\\cmidrule(r){3-6} \cmidrule(l){7-10} 	
		& & \multicolumn{2}{c}{MAP 50\%} & \multicolumn{2}{c}{MAP 25\%} & \multicolumn{2}{c}{MAP 50\%} & \multicolumn{2}{c}{MAP 25\%} \\ \cmidrule(r){3-4}\cmidrule(lr){5-6}\cmidrule(lr){7-8}\cmidrule(lr){9-10}
		Method & Embedding & QbE & QbS & QbE & QbS & QbE & QbS & QbE & QbS \\
		\thickhline
		Ctrl-F-Net & DCToW & 75.3 & 78.0 & 90.9 & 94.9 & 74.9 & 79.9 & 96.7 & 97.9 \\
		Ctrl-F-Net & PHOC & 73.9 & 77.2 & 91.0 & 94.5 & 67.9 & 74.7 & 95.8 & 97.8 \\
		Ctrl-F-Mini & DCToW & \textbf{81.2} & \textbf{83.8} & \textbf{92.1} & \textbf{95.2} & 86.2 & \textbf{90.4} & \textbf{97.0} & \textbf{98.3} \\
		Ctrl-F-Mini & PHOC & 79.1 & 81.5 & 90.3 & 93.9 & 85.7 & 89.1 & 95.8 & 97.9 \\
		
		TAU* \cite{pratikakis2016icfhr2016} & n/a & 37.5 & - & - & - & 61.8 & - & - & -\\	
		$\textnormal{LRC+SIFT}_{\textnormal{quant}}$ \cite{rothacker2017word} & PHOC & 74.5 & 78.8 & 80.4 & 85.3 & \textbf{91.1} & 89.9 & 95.6 & 95.3\\
		$\textnormal{AAM+SIFT}_{\textnormal{quant}}$ \cite{rothacker2017word} & PHOC & 69.4 & 74.0 & 75.9 & 80.3 & 89.6 & 88.9 & 96.2 & 96.0\\
		Encoder-Decoder Net \cite{axler2018toward} & PHOC & - & 79.0 & - & 78.7 & - & - & - & -\\
		\thickhline
	\end{tabular}
\end{table*}

\subsubsection{Computational and Storage Requirements}
We have computed inference and search times as well as storage requirements for the pre-processed images in Table \ref{tab:query_time} for all datasets. The inference and search times are averaged over pages and queries respectively. Across the board we see that the Ctrl-F-Net has slower inference times than the Ctrl-F-Mini, but surprisingly enough the search times are lower, even when storage space is larger (that is, more proposals to search through). This is most likely due to the non-max suppression step in the querying removing more proposals for the Ctrl-F-Net, which leads to fewer proposals to sort. The same process is most likely in effect in the cases when space requirements of the Ctrl-F-Mini are higher than the Ctrl-F-Net even with one only source of region proposals. In this case, the wordness score based non-max suppression removes enough DTP proposals using RPN proposals that the total number of proposals go down.

\subsection{State of the art comparison} \label{sec:sota}
In this section, we compare the best performing models determined from the ablation study in the previous section to the state-of-the-art in segmentation-free word spotting for all four public datasets and segmentation-based word spotting for the GW and IAM datasets. We adopt the Cosine loss instead of the Cosine Embedding loss, and for Ctrl-F-Net we make use of DTP proposals during training.

Table \ref{tab:sota} shows the results of the top Ctrl-F-Net and Ctrl-F-Mini models with the DCToW and PHOC embeddings, and contrasts them with the state of the art on the GW and IAM datasets. For the GW dataset, we outperform the other methods by a large margin, in both QbE and QbS and across the 25\% and 50\% overlap thresholds. The difference in MAP score between the GW 5-15 and GW 15-5 datasets are relatively small, considering the number of training and validation pages (5 vs 15) and test pages (15 vs 5). Our results that are marked with an asterisk use the evaluation protocol of \cite{Ghosh_word_spotting, ghosh2015query, ghosh2018text}. A small caveat is needed for the Washington dataset for its annotation quality. As the dataset was manually annotated, there can be a relatively large amount of empty space around the words. This has little effect for longer words, but for single letter words, this can cause perfectly segmented words to have less than 50\% overlap with the ground truth. Although, the issue is much smaller for the 25\% overlap threshold. 

For the IAM dataset, the recently introduced model in \cite{axler2018toward} has very high performance, beating the previous version of this work. However, the changes introduced in this paper led us to outperform their model. Since the first version of this paper, the Ctrl-F-Mini has been adopted in \cite{krishnan2019hwnet} with their own learned embedding to achieve the highest mAP for QbE on IAM. We note that the results in \cite{Ghosh_word_spotting} and \cite{ghosh2015query} on the IAM dataset is not directly comparable, as they do line spotting where whole lines are retrieved and they perform their search in the annotated text lines, not the full pages, and their distance between a query and a text line is the shortest distance between the query and the word candidates of that line. According to the results presented in \cite{almazan2014word}, this is a slightly easier task.

In Table \ref{tab:sota_konz_botany} we compare our models on the Botany and Konzilsprotokolle datasets and note that in all but one setting (QbE 50\% overlap) we outperform existing methods, often with a large margin. For example, with QbS using 25\% overlap on the Botany dataset we improve with over 10 percentage points, reducing the error by 67\%. We can observe a similar result for Konzilsprotokolle, where the error is reduced by 68\%.	

In Table \ref{tab:sota_seg_comp}, we compare the best segmentation-free setup with a 25\% overlap threshold with state of the art methods for segmentation-based word spotting, that use the same evaluation protocol. We observe that we have competitive results for the GW 15-5 split in both QbE and QbS, whereas for IAM the QbS performance is quite close the best segmentation-based methods, even though they depend on manually segmented bounding boxes. We also include some of the top methods for line-level word spotting for the IAM dataset as it is reasonably easy to segment into lines. While the numbers are not as directly comparable as the word-level methods ( due to a different query set and vocabulary, additional data for language models, retrieval of lines rather than words to name a few), we include them for additional context.

\begin{table}
	\renewcommand{\arraystretch}{1.3}
	\caption{MAP comparison in \% with state-of-the-art segmentation-based methods using a 25\% overlap threshold and the GW 15-5 split. Methods marked with $^\dagger$ use ground truth segmented word images, but otherwise use a similar evaluation method. 
	}\label{tab:sota_seg_comp}
	\centering
	\begin{tabular}{lcccc}

		&\multicolumn{2}{c}{GW} & \multicolumn{2}{c}{IAM} \\ \cmidrule(lr){2-3}  \cmidrule(lr){4-5} 
		Method & QbE & QbS & QbE & QbS \\
		\thickhline
		Ctrl-F-Net PHOC & 97.3 & 96.4 & 78.0 & 88.4 \\
		Embed attributes$^\dagger$ \cite{almazan2014word} &  93.0 & 91.3 & 55.7 & 73.7 \\
		DCToW$^\dagger$ \cite{wilkinson2016semantic} & 98.0 & 93.7 & 77.0 & 85.3 \\
		TPP-PHOCNet (CPS)$^\dagger$ \cite{sudholt2017attribute} &  98.0 & 97.9 & 82.7 & 93.4 \\
		DeepEmbed$^\dagger$ \cite{krishnan2018word} &  98.0 & \textbf{98.9} & 90.4 & \textbf{94.0} \\
		Hwnet v2$^\dagger$ \cite{krishnan2019hwnet} &  \textbf{98.2} & - & \textbf{92.4} & - \\		
		\thickhline
	\end{tabular}
\end{table}

\subsection{Discussion}
We make several observations from the experiments. The first is that we get an increase in the MAP score across the board from lowering the overlap threshold to 25\%. This suggests that there is further performance to be gained from more accurate proposals. Another possible explanation is that this effect is due to tightness of the annotation for a dataset. This means that for single letter words like "I", the amount of space surrounding the ground truth box makes a region proposal that tightly attends to the ink overlapping less than 50\%. This effect is greater for manually labelled datasets like the GW datasets, and less so for the IAM dataset, which is reflected in the results. However, for the application of word search in manuscripts as a way to assist scholars (as considered in the following Section), 25\% overlap is more than enough as the user would in any case manually inspect each result.

The results in Table \ref{tab:sota_seg_comp} show a minor difference in performance between the segmentation-free and segmentation-based word spotting methods. The segmentation-based method most similar to this work is \cite{wilkinson2016semantic}, and the overall performance of the Ctrl-F-Net is higher. This suggests that there might be possible upsides to using a segmentation-free approach when learning a representation for a word image. For example, the increased consistency of automatic region proposal methods compared to manual labeling could be beneficial for learning.

A result that is of great significance for practical adoption of this work in historical research, is the high performance on GW 5-15 dataset, where we train on 4 pages, validate on 1 page and test on 15 pages. This suggests that the model is learning efficiently with respect to the amount of training data. This is crucial when working with historical manuscripts, which are very expensive to annotate due to the expert knowledge required. An important aspect of training on such small data is extensive use of data augmentation. To that end we have introduced two complementary data augmentation techniques for full page manuscripts to facilitate the learning of proposed models. 

From the experiments presented in the previous section, we conclude that using the RPN as a main source of region proposals without any external region proposals is a suboptimal approach. While there has been some work adapting the RPN for text proposals \cite{liao2018textboxes++}, two of the suggested improvements relate to handling arbitrarily oriented text and increasing recall. Both of these are not issues here, in fact, the RPN has higher recall than the DTP. A third adaptation of the RPN is changing the aspect ratio of the anchor boxes, which has been adopted our formulation. Instead, the results suggests that the RPN, while generic in its current form, could be improved upon or specialized for sub-domains of computer vision. One example relevant for this work would be how to generate proposals that work well for word spotting.

Compared to the natural image object detectors (Multibox \cite{erhan2014scalable} and YOLO \cite{redmon2016you}) that were evaluated in \cite{moysset2018learning}, the RPN is a capable source of region proposals for manuscript images. Because of its implementation as a convolution (i.e., sliding window) where a set of anchor boxes are regressed into every position on the feature map, it is able to detect hundreds of objects per image. While YOLO fails to work at all for manuscripts in \cite{moysset2018learning}, Multibox does reasonably well but is outperformed by their proposed approach. A possible reason as to why the RPN works well in terms of recall is because to the large amount of initial proposals that are then reduced via score thresholding and non-max suppression.

\begin{figure}[t!]
	\begin{center}
		\includegraphics[width=0.99\linewidth]{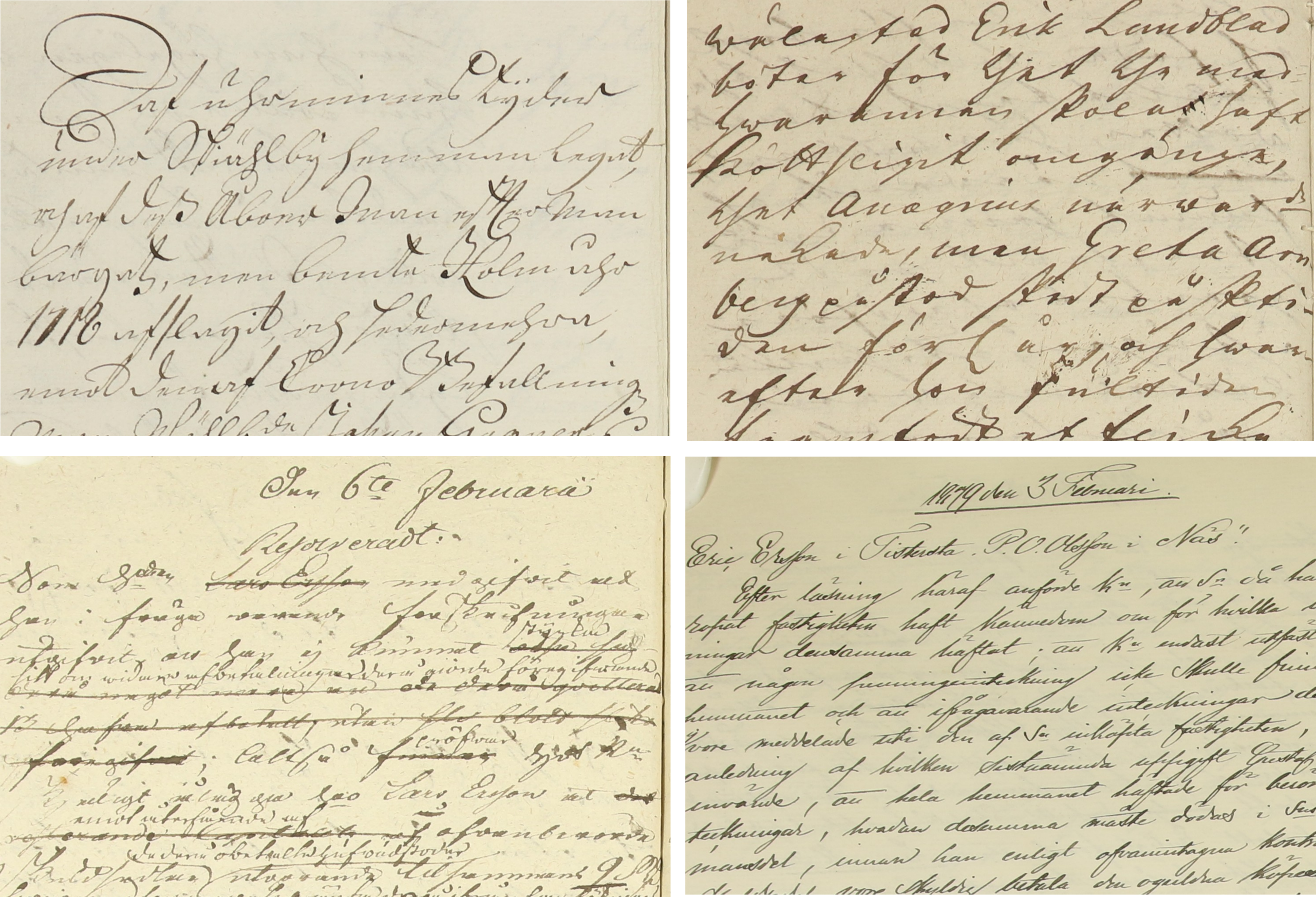}
	\end{center}
	\caption{Parts of 4 sample images drawn from the Snevringe dataset. Going from top left to bottom right, they are written in 1719, 1735, 1797, and 1879.}
	\label{fig:dombok}
\end{figure}

Furthermore, the ablation study showed how the Ctrl-F-Mini performed on par with the full Ctrl-F-Net, and in some instances even outperformed the full model. This held true also in the small data setting of GW 5-15. Considering the reduced model complexity, and decreased training and inference times Ctrl-F-Mini provides over Ctrl-F-Net, it is a recommended alternative to the full model when it comes to easily segmented data. However, the Ctrl-F-Mini with DTP proposals still suffers from the known limitations of connected components based methods for segmenting. So for more densely written, noisy manuscripts where document binarization is difficult and large parts of the text is one connected component, the full Ctrl-F-Net with its sliding window RPN would be recommended.

\section{Case study: early modern court records}\label{sec:case_study}
Court records are often used in historical research. Their usefulness stretches far beyond the study of crime and judicial systems as they offer insights into practices and mentalities of ordinary people that very few sources do. In the words of a well-known introduction to historical methodology, "court records are probably the single most important source we have for social history of the medieval and early modern periods." \cite{tosh2015pursuit}. Not surprisingly, several of the most famous and influential historical studies are based on court record \cite{ladurie2013montaillou, ginzburg1992cheese, zemon1983return}. At the same time, the richness in information and the variety of subjects dealt with in a single volume make the records difficult to work with. Swedish court records mix criminal and civil cases and they often lack even simple search tools such as indexes. Finding relevant information is extremely time-consuming and researchers need often to restrict their empirical research to a very limited number of volumes.

\textbf{The Snevringe Court Records} consists of 64 volumes of newly digitized court records from the magistrate court of Snevringe judicial district, written between 1719 and 1880. The 64 volumes consist of 55k images, each of which contain 2 pages. Figure \ref{fig:dombok} shows a sample of four images of the dataset. The court records provide several challenges: unlike modern text, there is no standardized spelling; said spelling evolves over time, compounding the problem; and there are hundreds of different writers, adding their personal variation. A final peculiarity of the court records is that they are written during a time in Sweden where a change of script took place. Earlier volumes use Kurrent script (or German cursive), with particular words written in Latin cursive. A gradual, non-linear change of scripts to Latin cursive occurs over the time span that the court records were written. This dual script provides interesting challenges where certain characters are written in two completely different ways in the same dataset. The queries we evaluate are chosen according to their relevance to contemporary historical research.


We have manually annotated 11 pages for training, where 3 are from the Snevringe set of court records (these pages are removed when searching). The rest are from another set of court records from adjacent judicial districts and a nearby town. We use the Ctrl-F-Net and training is done as detailed in section \ref{sec:training} except that initialized with a model trained on the IAM dataset.

\begin{table}
	\renewcommand{\arraystretch}{1.3}
	\caption{Quantitative results for the queries used for the Snevringe dataset. $P(k)$ is the precision at rank k. OOV denotes out-of-vocabulary, meaning not present in training vocabulary.	}
	\label{tab:snevringe}
	\centering
	\begin{tabular}{lccccc}
		Queries & Translation & OOV & P(1) & P(10) & P(50)\\
		\hline
		str{\"o}msholm & Str{\"o}msholm & & 1.00 & 1.00 & 0.98 \\
		wester{\aa}s & V{\"a}ster{\aa}s & & 1.00 & 1.00 & 1.00 \\
		madame & madame & & 1.00 & 0.70 & 0.20 \\
		stalldr{\"a}ng & stableman &  & 1.00 & 0.70 & 0.52 \\
		vidk{\"a}ndt & acknowledged  &  & 1.00 & 0.80 & 0.58 \\		
		l{\"a}nsmannen & county sheriff  & & 1.00 & 1.00 & 0.8 \\
		februari & February & \checkmark & 1.00 & 0.80 & 0.58 \\
		informator & tutor & \checkmark & 0.00 & 0.10 & 0.02 \\
		gifta & married & \checkmark & 0.00 & 0.70 & 0.48 \\
		sala & Sala & \checkmark & 0.00 & 0.00 & 0.04 \\
		\hline
		Average & n/a &  & 0.70 & 0.68 & 0.52 \\
	\end{tabular}
\end{table}

Because we are working with an unexplored collection of manuscripts, doing an exhaustive evaluation is not possible. Instead we adopt common web-scale metrics that do not require knowledge of $R$, the number of relevant instances for a query. We manually annotate the top-50 results for each query and calculate the $P_k$ for each query at different at $k=\{1, 10, 50\}$. Contrary to the MAP, this metric has no measure of the ordering of results, only the precision of the top-k results. 

\begin{figure}
	\begin{center}
		\includegraphics[height=0.78\linewidth]{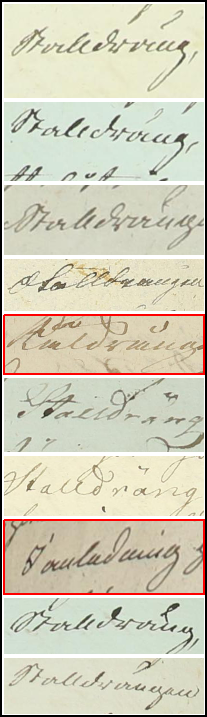}
		\includegraphics[height=0.78\linewidth]{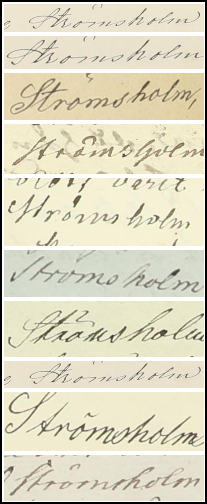}
		\includegraphics[height=0.78\linewidth]{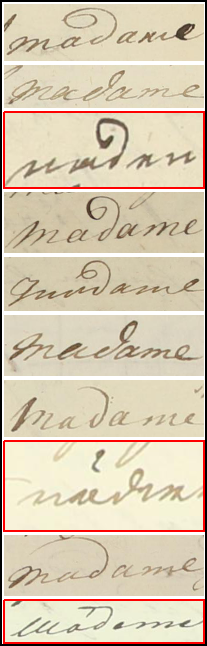}
		\includegraphics[height=0.78\linewidth]{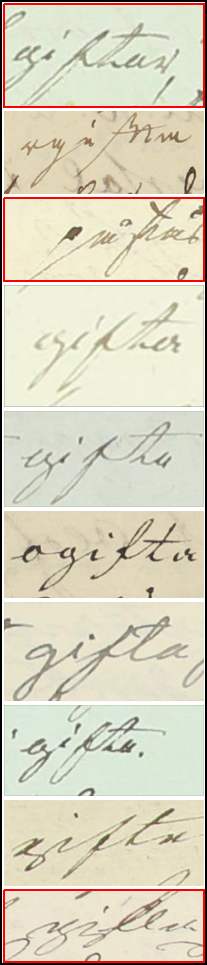}
	\end{center}
	\caption{Qualitative search results. The figure depicts the top 10 results starting from the top for the four queries "Stalldr{\"a}ng" (stableman), "Str{\"o}msholm", "madame", and "gifta" (married). Incorrect retrievals are highlighted in red.}
	\label{fig:qualitative_results}
\end{figure}

Quantitative results are presented in Table \ref{tab:snevringe}. We show the performance of 10 queries relevant for contemporary historical research. As expected, the words that are present in the training vocabulary perform the best on average. The out-of-vocabulary (OOV) words seem to perform worse, which is not so surprising as we are doing zero-shot retrieval. An interesting exception is the query "gifta" (eng. married), for which the results were surprisingly good. Another noteworthy aspect for the query "informator" (eng. tutor) is that while it is most commonly written using Latin script, the model seems to be searching for the query using Kurrent script, illustrating some of the difficulties in working with this data. This is likely due to the Kurrent script being heavily overrepresented in the training data. We further provide qualitative results in Figure \ref{fig:qualitative_results} to showcase some variability of the writing styles used in the court records.

With the Snevringe court records we provide results on a set of queries being investigated in contemporary historical research. In essence, we are testing our model in the wild, i.e., directly evaluating our proposed approach in a setting where it is designed to be deployed. The data is unexplored, un-curated and noisy. There are blank pages, images of book covers, and extremely messy pages with lots of stricken out text, faded ink, and extensive notes between text lines. This, together with its size, makes the court records more difficult to work with than any publicly available word spotting dataset. 

The value of word spotting for research in historical manuscripts cannot be overstated. The work often entails manually locating small pieces of information scattered throughout large amounts of texts, and finding only where to look can be very time consuming. In effect, limited sets of elusive data are what a historians' interpretations are based on, and a limiting factor when it comes to which inquiries can be conducted at all. Speeding up the process of identifying relevant sections in handwritten texts would not only make it possible to gather more data, but also make way for new questions to be researched.

\section{Conclusion}
We have introduced Ctrl-F-Net, a model for segmentation-free query-by-string word spotting. It simultaneously produces region proposals, and embeds them into a word embedding space in which searches are performed. Using an ablation study, we investigate several model choices, most notably the source of region proposals. The ablation leads us to propose the simplified Ctrl-F-Mini, a model suited to manuscripts that easily segmented into words. Our models outperform the previous state-of-the-art approaches for segmentation-free word spotting, in some cases by a large margin. Moreover, in a case study applying the Ctrl-F-Net to a collection of 64 volumes of court records, spanning the most of the 18\textsuperscript{th} and 19\textsuperscript{th} centuries, we enable a historical study using orders of magnitude greater data than would be possible otherwise. 

\ifCLASSOPTIONcompsoc
  \section*{Acknowledgments}
  This project is a part of q2b, From quill to bytes, which is a digital humanities initiative sponsored by the Swedish Research Council (Dnr 2012-5743), Riksbankens Jubileumsfond (NHS14-2068:1) and Uppsala university.
\else
  \section*{Acknowledgment}
\fi

\ifCLASSOPTIONcaptionsoff
  \newpage
\fi



%
%
%

\bibliographystyle{IEEEtran}

\bibliography{neural_search}

%


\vfill


\end{document}